# PRISM-Net: Dynamic Routing Network with VLM-Guided Disentangled Attribute Distillation for Explainable Fine-Grained Product Retrieval

Chung-Ming Lo*, *Member, IEEE*, Che-Wei Chang

***Abstract*—Fine-grained visual product retrieval in large-scale e-commerce multimedia systems demands discriminative representations capable of simultaneously resolving macro-level category semantics and micro-level visual attributes such as structural contour (line) and chromatic appearance (color). However, conventional backbones including convolutional neural networks (CNNs) and vision Transformers (ViTs) inherently suffer from feature entanglement and shortcut learning. To learn the micro-level visual attributes without costly manual attribute annotations, we propose phase-routed interpretable semantic mixture-of-experts network (PRISM-Net), an explainable multi-expert architecture that disentangles holistic visual features into disentangled line and color expert pathways governed by a conditional meta-router. In phase 1, we introduce an external VLM-Guided sigmoid semantic distillation mechanism with gradient-detached routing gates, utilizing zero-shot text prompts from SigLIP to supervise specialized structural and chromatic experts. In phase 2, the disentangled expert embeddings are adaptively fused with macro base features via a learned meta-router gate to calibrate category boundaries. Extensive 5-fold cross-validation across 275,454 e-commerce images (18 categories) and full-scale retrieval benchmarks (55,091 queries against a 220,363-image gallery) across SwinV2, ConvNeXt-V2, and DINOv3 demonstrate substantial gains. On SwinV2, PRISM-Net boosts full-fold product mAP@10 from 80.98% to 83.90% (+2.92%), line mAP@10 by +5.74%, and color mAP@10 by +9.12%. Furthermore, unsupervised distribution analysis of the meta-router gate reveals that both ViT and CNN backbones emergently allocate up to 6.8x higher structural line weights to rigid appliances and hardware while dedicating >96% chromatic weight to clothing.**

## I. INTRODUCTION

The popularity of visual e-commerce and multimedia content platforms has made content-based image retrieval (CBIR) a cornerstone technology for modern digital retail [1], [2]. Unlike generic object recognition that merely assigns broad category labels, fine-grained product retrieval requires a vision system to simultaneously satisfy two hierarchical criteria: (i) macro-level semantic accuracy, ensuring that retrieved candidates belong to the exact product taxonomy, and (ii) micro-level perceptual alignment, ensuring that top-ranked candidates preserve tiny visual attributes such as geometric contour, structural contour, fabric texture, and chromatic appearance [3], [4]. In real-world long-tail catalogs containing hundreds of thousands of items, standard holistic representations extracted by convolutional neural networks (CNNs) [5], [6] or vision Transformers (ViTs) [7], [8] suffer from severe feature entanglement. Because standard cross-entropy training encourages shortcut learning [9], holistic embeddings frequently combine background illumination or dominant color blocks with category identity. Consequently, querying a metallic industrial tool often retrieves unrelated specular kitchenware, whereas querying a crimson garment frequently returns items of the correct clothing category but with wildly mismatched colors.

Recently, large-scale vision-language models (VLMs) such as CLIP [10], SigLIP [11], and EVA-CLIP [12] have transformed multimodal representation learning by aligning hundreds of millions of image-text pairs in a shared embedding space. Their linguistic priors provide zero-shot access to human-interpretable attributes (e.g., 'thick structural lines' or 'magenta hue'). Recent retrieval and adaptation studies therefore combine frozen VLM priors with task-specific inversion or knowledge-enhanced prompt adaptation rather than relying on unmodified embeddings [13], [14]. Text-aligned image encoders compress high-frequency geometry into coarse semantic representations, whereas domain-finetuned visual backbones preserve local discrimination but lack explicit controls for separating structural and chromatic evidence. Self-supervised foundation models such as DINOv2 [15] and DINOv3 [16] further strengthen pixel-level fidelity, yet still expose no mechanism for attribute-specific routing during similarity ranking. This mismatch motivates transferring VLM semantics as routing supervision rather than using the VLM embedding itself as the retrieval representation.

Conditional computation frameworks, most notably sparse and soft mixture-of-experts (MoE) [17], [18], [19], offer scalable model capacity. Recent multimodal MoE systems have begun separating semantic and structural expert branches [20], but their routing remains optimized primarily for

This work was supported in part by the National Science and Technology Council (NSTC 112-2221-E-A49 -188 -MY3) (*Corresponding author:* Chung-Ming Lo)

C.M. Lo is with the Institute of Artificial Intelligence Innovation, Industry Academia Innovation School, National Yang Ming Chiao Tung University, Hsinchu, Taiwan (e-mail: buddylo@nycu.edu.tw).

C.W. Chang is with the Graduate Institute of Library, Information and Archival Studies, National Chengchi University, Taipei, Taiwan (email: google86929@icloud.com)

downstream recognition rather than explicit attribute disentanglement. Such routing neither guarantees line–color specialization nor prevents experts from encoding redundant mixtures of shape and appearance. The unresolved technical challenge is therefore to convert external linguistic priors into disentangled, attribute-specific routing constraints while preserving the native spatial precision of heterogeneous visual backbones and avoiding manual attribute annotation.

This study proposes phase-routed interpretable semantic mixture-of-experts network (PRISM-Net) to address this challenge through a phase-separated semantic routing architecture rather than conventional task-only MoE gating. Analogous to an optical prism separating a composite beam, the network decomposes entangled backbone features into a 3-expert line pathway for structural contours and a 10-expert color pathway for chromatic appearance, while preserving a global semantic shortcut for category identity. In phase 1, frozen SigLIP text embeddings are converted into independent Sigmoid supervision signals for the two routers; gradient-detached modulation isolates router guidance from expert feature learning and suppresses collapse or semantic leakage between pathways. An instance-adaptive meta-router then estimates gate alpha in [0, 1] to conditionally fuse structural, chromatic, and global evidence. In phase 2, freezing the disentangled representation before classifier calibration decouples attribute formation from decision-boundary optimization, thereby avoiding the gradient competition inherent in joint multi-objective training.

The contributions are summarized in three aspects:

- Semantically Constrained Disentangled Expert Decomposition: Unlike conventional visual MoE architectures that discover unconstrained and often redundant experts, PRISM-Net assigns non-overlapping structural and chromatic semantics to dedicated line and color expert groups. A parallel global shortcut preserves macro-level category information, producing a hierarchical representation that jointly models identity, contour, and appearance without collapsing them into a single holistic embedding.
- VLM-to-Router Semantic Distillation with Gradient Isolation: Frozen SigLIP priors are repurposed as attribute-level routing targets instead of retrieval features or holistic teacher embeddings. Independent Sigmoid supervision permits multiple visual attributes to coexist, while stop-gradient modulation prevents teacher constraints from overwriting backbone representations and blocks mutual interference between routing decisions and expert feature optimization.
- Instance-Adaptive Meta-Fusion and Phase-Decoupled Optimization: A conditional meta-router learns image-dependent structural–chromatic weighting rather than applying fixed feature concatenation or globally shared expert coefficients. The two-phase schedule further separates semantic disentanglement from classifier calibration, reducing multi-task gradient conflict and allowing the same routing principle to operate across CNN, hierarchical ViT, and self-supervised foundation backbones.

## II. RELATED WORK

Fine-grained image retrieval aims to differentiate subcategories or specific product instances that share high visual similarity [1], [3]. Beyond retail applications, multilevel CNN representations have supported large-scale hierarchical medical image retrieval [21], while complementary deep-image and vision-language embeddings have improved artery-segment recognition in carotid ultrasonography [22]. Journal studies have investigated identity-preserving latent attributes [23], global–local multimodal features [24], and part-aware representations [25] to improve fine-grained discrimination. However, such approaches often assume person-centric structure or require domain-specific supervision, limiting their direct application to multi-category e-commerce catalogs where geometry and color co-vary across arbitrary viewpoints. Prior studies have also explored attribute-guided retrieval and design-attribute disentanglement [4], [26] through supervised representation learning. However, collecting dense ground-truth annotations for contour thickness, surface geometry, and multi-color palettes across hundreds of thousands of products is prohibitively expensive. In contrast, the proposed PRISM-Net eliminates manual attribute labeling entirely by using disentangled soft attribute distributions from pre-trained VLMs.

Conditional computation via MoE [17] has become a common approach for scaling foundation models. In computer vision, V-MoE [18] and Soft-MoE [19] route image patch tokens across expert MLPs to expand parameter capacity while maintaining manageable FLOPs. More recently, multimodal expert routing [20] and parameter-efficient adaptation methods such as LoRA [27] have broadened expert specialization and efficient fine-tuning. Despite their computational elegance, conventional visual MoE routers are optimized mainly by downstream task losses and load-balancing regularizers. Consequently, specialization often remains difficult to interpret and may collapse into redundant feature extractors in fine-grained retrieval. PRISM-Net fundamentally departs from standard MoE by binding each expert group to an explicit physical attribute space (line vs. color) via cross-modal distillation and introducing a second-stage conditional meta-router that learns category-specific attribute weighting.

Contrastive vision-language models (CLIP [10], SigLIP [11]) and self-supervised visual foundation models (DINOv2 [15], DINOv3 [16]) represent two distinct poles of modern visual representation learning. Recent journal studies use knowledge-enhanced prompting [14], dynamic contrastive distillation [28], multilingual vision-language distillation and alignment [29], and cross-modal adapters [30] to adapt or transfer VLM knowledge to downstream representations. Multimodal long-context models and unified mixtures of experts further demonstrate complementary semantic reasoning and expert integration [31], [32]. However, two critical gaps remain unaddressed in the literature. First, prior adaptation and distillation methods generally preserve a holistic representation, thereby inheriting the VLM's linguistic compression bottleneck and losing fine-grained structural cues. Second, existing studies rarely investigate how heterogeneous student architectures (pure CNNs vs. hierarchical ViTs vs. self-supervised ViTs) interact with VLM teachers when student visual capacity approaches or exceeds that of the teacher. Accordingly, the experiments evaluate different backbone families under a shared SigLIP teacher to clarify how architectural compatibility and student capacity shape the resulting distillation relationship.

## III. MATERIALS AND METHODS

### *A. Amazon Product Dataset*

The product image data used in this study were obtained from the Amazon Products Dataset available on Kaggle (https://www.kaggle.com/datasets/lokeshparab/amazon-products-dataset). After data cleaning, categories containing fewer than 100 images were excluded, and semantically similar categories were merged. This category filtering process retained 18 major product categories, yielding a total of N = 275,454 high-resolution images for the subsequent classification and retrieval experiments. Unlike synthetic or single-domain benchmarks (e.g., clothing-only or bird-only datasets), this benchmark spans both rigid structural goods (appliances, industrial supplies, car & motorbike parts, musical instruments, audio/camera equipment) and non-rigid chromatic goods (men's and women's clothing, footwear, beauty & health, grocery & gourmet foods).

Especially, the dataset exhibits a realistic natural long-tail distribution [33], where sample volumes vary by two orders of magnitude from head categories such as Accessories (60,145 images) and Women's Clothing (39,070 images) down to tail categories such as Pet Supplies (1,205 images) and Music (735 images). To describe the class composition and long-tail distribution without relying on experimental outcomes, the 18 categories are partitioned into three mutually exclusive subsets according to sample volume: (1) head classes (top 6 by volume, accounting for 79.4% of total samples); (2) medium classes (middle 6 by volume, 15.5% of samples); and (3) tail classes (bottom 6 by volume, 5.1% of samples). Table I details the categories included in each subset, their sample distribution, and their dominant visual characteristics.

**TABLE I**
LARGE-SCALE E-COMMERCE MULTI-ATTRIBUTE BENCHMARK TAXONOMY AND LONG-TAIL DISTRIBUTION (275,454 IMAGES)

| **Subset Tier** | **Product Categories Included (6 Classes per Tier)** | **Total Images** | **Fold 0 Query (20%)** | **4-Fold Gallery (80%)** | **Dominant Visual Cue** |
|---|---|---|---|---|---|
| Head Classes (Top 6) | Accessories, TV/Audio/Cameras, Women's Clothing, Men's Shoes, Men's Clothing, Appliances | 218,810 (79.4%) | 43,762 | 175,048 | Mixed Chromatic & Rigid |
| Medium Classes (Mid 6) | Home & Kitchen, Kids' Fashion, Sports & Fitness, Beauty & Health, Bags & Luggage, Car & Motorbike | 42,770 (15.5%) | 8,554 | 34,216 | Complex Multi-Part / Texture |
| Tail Classes (Bot 6) | Toys & Baby, Women's Shoes, Industrial Supplies, Grocery & Gourmet, Pet Supplies, Music | 13,874 (5.1%) | 2,775 | 11,099 | Fine Contour & Color Nuance |
| Total / Protocol | All 18 Commercial Product Categories (5-Fold Cross-Validation, Seed = 42) | 275,454 (100%) | 55,091 Queries | 220,363 Gallery | 3 Line + 10 Color Priors |

### *B. PRISM-Net*

Fig. 1 summarizes the data organization and the end-to-end classification and retrieval workflow. The 275,454 product images from 18 categories are organized under a five-fold protocol, with one fold used as the query set (55,091 images; 20%) and the remaining four folds used as the gallery set (220,363 images; 80%). During offline training, the catalog images and frozen SigLIP text priors enter phase 1, where the backbone representation is disentangled into three line experts and ten color experts; stop-gradient routing prevents the classification objective from overriding attribute specialization. In phase 2, an adaptive meta-router learns an image-dependent gate α ∈ [0, 1] to balance structural and chromatic evidence and calibrate the category representation. The resulting gallery embeddings are cached as an FP16 GPU feature table. During online retrieval, a query image is encoded through the same structural and chromatic pathways, fused into a query embedding, and compared with the gallery by cosine similarity to return a real-time ranked list that preserves both category consistency and fine-grained line and color similarity.

Fig. 2 illustrates how PRISM-Net produces a shared representation for product classification and large-scale image retrieval. The vision backbone first encodes the input image as a feature F and a global base feature $f_{base}$. The global branch retains category-level semantics, while the other branch are processed by two attribute-specific pathways during phase 1. The line pathway contains three residual experts for geometric contour modeling, whereas the color pathway contains ten chromatic residual experts for palette modeling. Their routers are distilled from a frozen SigLIP teacher and stop-gradient task routing prevents the downstream category objective from disrupting expert specialization.

Given an input RGB product image X in $R^{(H \times W \times 3)}$, a visual backbone extracts a spatial feature map F in $R^{(C_b \times H' \times W')}$, where $C_b$ denotes the channel dimension. A macro-level global representation $f_{base}$ in $R^{(C_b)}$ is simultaneously obtained via spatial global average pooling:

$$f_{base} = \frac{1}{H'W'} \sum F_{h,w}, \quad f_{base} \in \mathrm{R}C^{b} \quad (1)$$

In phase 2, a meta-router estimates an instance-dependent coefficient α ∈ [0, 1] from the line and color features and dynamically fuses the two attribute representations into a routed feature. This feature is concatenated with $f_{base}$ to form a hybrid representation. Consequently, classification enforces macro-level category consistency, whereas gallery ranking captures fine-grained structural and chromatic similarity within the predicted product space.

### 1) Disentangled Line and Color Expert Routing

To decouple geometric structure from chromatic appearance, $F$ is dispatched in parallel to a line pathway (comprising $M = 3$ structural experts $\{E_L^1, E_L^2, E_L^3\}$ corresponding to thick, thin, and medium contour profiles) and a color pathway (comprising $N = 10$ chromatic experts $\{E_C^1, ..., E_C^{10}\}$ spanning ten canonical color subspaces). Each expert $E_L^m$ and $E_C^n$ is instantiated as a dedicated residual convolutional bottleneck block with batch normalization and ReLU activation. The line router $R_L$ processes $F$ via an adaptive pooling and MLP projection head to predict unnormalized structural logits $z_L$ in $R^3$, whereas the color router $R_C$ projects $F$ to predict chromatic logits $z_C$ in $R^{10}$. Rather than enforcing mutually exclusive Softmax competition, which erroneously assumes a product cannot exhibit multiple color tones or mixed contour scales, non-exclusive Sigmoid gating probabilities are formulated:

$$p_L = \sigma(R_L(F)) \in [0, 1]^3, \quad p_C = \sigma(R_C(\mathrm{F})) \in [0, 1]^{10} \quad (2)$$

During joint training, backpropagating downstream product classification gradients directly through $p_L$ and $p_C$ into the routers can corrupt the VLM-aligned semantic meaning of the gates, causing the routers to degenerate into arbitrary feature re-weighters. To prevent this phenomenon, a router gradient protection operator StopGrad(.) is introduced via tensor detachment, which isolates the router parameters from classification noise so they are updated exclusively by the external VLM distillation signal. The aggregated spatial feature maps $F_{line}$ and $F_{color}$ are computed as:

$$F_{line} = \sum_{\mathrm{m}=1}^{M} \tilde{p}_{L,m} \cdot E_L^m(F), \quad \tilde{p}_{L,m} = \mathrm{StopGrad}(p_{L,m}) \quad (3)$$

$$F_{color} = \sum_{\mathrm{n}=1}^{N} \tilde{p}_{C,n} \cdot E_C^n(F), \quad \tilde{p}_{C,n} = \mathrm{StopGrad}(p_{C,n}) \quad (4)$$

### 2) VLM-Guided Sigmoid Semantic Distillation

To supervise the line and color routers without human annotations, a frozen vision-language foundation model, SigLIP-base-patch16-224 [11], is employed as an external semantic teacher. Two attribute-specific prompt banks are constructed: a structural prompt bank $T_L$ = {'a product with thick lines', 'a product with thin lines', 'a product with medium lines'} and a chromatic prompt bank $T_C$ = {'a black product', 'a blue product', 'a cyan product', 'a dark gray product', 'a green product', 'a light gray product', 'a magenta product', 'a red product', 'a white product', 'a yellow product'}. Let $v_{img}$ in $R^d$ and $\{t_L^m\}$, $\{t_C^n\}$ in $R^d$ denote the L2-normalized image and text embeddings extracted by the frozen VLM teacher. The teacher's zero-shot soft probability targets $q_L$ in $[0, 1]^3$ and $q_C$ in $[0, 1]^{10}$ are generated via:

$$q_{L,m} = \sigma(\tau \cdot \langle v_{img}, t_L^m \rangle + \mathrm{b}), \quad q_{C,n} = \sigma(\tau \cdot \langle v_{img}, t_C^n \rangle + \mathrm{b}) \quad (5)$$

where tau and b denote the learned logit scale and bias of SigLIP. The student routers' predicted probability distributions $p_L$ and $p_C$ are aligned with the VLM teacher's soft targets $q_L$ and $q_C$ using point-wise binary cross-entropy (BCE) distillation losses:

$$\mathcal{L}_{distill}^{L} = -\frac{1}{3} \sum [q_{L,m} \log(p_{L,m}) + (1 - q_{L,m}) \log(1 - p_{L,m})] \quad (6)$$

$$\mathcal{L}_{distill}^{C} = -\frac{1}{10} \sum [q_{C,n} \log(p_{C,n}) + (1 - q_{C,n}) \log(1 - p_{C,n})] \quad (7)$$

### 3) Conditional Meta-Router Gate

Because different product categories rely asymmetrically on geometric contour versus chromatic appearance. A conditional meta-router $M_{meta}$, parameterized by a two-layer bottleneck MLP with intermediate hidden dimension 64, is introduced. Given the concatenated specialist vectors $[f_{line} \,||\, f_{color}]$ in $R^{(2C_b)}$, $M_{meta}$ predicts a scalar instance-specific routing gate alpha in [0, 1]:

$$\alpha = \sigma(W_2 \cdot \mathrm{ReLU}(W_1[f_{line} \parallel f_{color}])) \in [0, 1] \quad (8)$$

The disentangled micro-attribute embedding $f_{routed}$ is then convexly combined via alpha and concatenated with the macro global shortcut $f_{base}$ to construct the final L2-normalized hybrid retrieval representation $f_{final}$ in $\mathrm{R}^{(2C_b)}$:

$$f_{routed} = \alpha \cdot f_{line} + (1 - \alpha) \cdot f_{color} \quad (9)$$

$$f_{final} = \frac{[f_{base} \parallel f_{routed}]}{\|[f_{base} \parallel f_{routed}]\|_2} \in R^{2C_b} \quad (10)$$

### 4) Two-Phase Optimization and Multimodal Ranking

Phase 1 use the first 70% of training epochs, $T_1 = 21$ for 30-epoch schedules, the visual backbone, routers, experts, and classifier are jointly optimized using a composite objective combining label-smoothed cross-entropy loss $\mathcal{L}_{\mathrm{CE}}$ (smoothing factor $\varepsilon = 0.1$) and the VLM distillation losses:

$$\mathcal{L}_{\mathrm{Phase1}} = \lambda_{\mathrm{prod}}\, \mathcal{L}_{\mathrm{CE}}(W_{\mathrm{cls}}\, f_{\mathrm{final}}, \mathrm{y}) + \lambda_{\mathrm{L}}\, \mathcal{L}_{\mathrm{distill}}^{\mathrm{L}} + \lambda_{\mathrm{C}}\, \mathcal{L}_{\mathrm{distill}}^{\mathrm{C}} \quad (11)$$

In Phase 2 (the final 30% of epochs, epochs 22 to 30), the visual backbone, line/color routers, and expert branches are frozen to preserve the disentangled attribute manifolds. Optimization is dedicated exclusively to training the meta-router and calibrating the final classification and metric boundary ($\lambda_{\mathrm{prod}} = 1.0$). During retrieval evaluation, given a query embedding $q = f_{\mathrm{final}}(X_{\mathrm{q}})$ and a gallery database $\mathcal{G} = \{g_{\mathrm{j}}\}_{\mathrm{j}=1}^{N_{\mathrm{g}}}$, candidates are ranked by cosine similarity $\mathrm{S}(\mathrm{q}, g_{\mathrm{j}}) = q^{\mathrm{T}} g_{\mathrm{j}}$. Beyond standard binary mean average precision (mAP@K), perceptual ranking alignment is evaluated via normalized discounted cumulative gain (NDCG@K):

$$\mathrm{NDCG@K} = \frac{\mathrm{DCG@K}}{\mathrm{IDCG@K}}, \quad \mathrm{DCG@K} = \sum_{\mathrm{r}=1}^{\mathrm{K}} \frac{rel_{\mathrm{r}}}{log_2(\mathrm{r}+1)} \quad (12)$$

where $rel_{\mathrm{r}} \in [0, 1]$ represents the semantic relevance score at rank $r$, and IDCG@K is the ideal DCG metric obtained by sorting relevance scores in descending order.

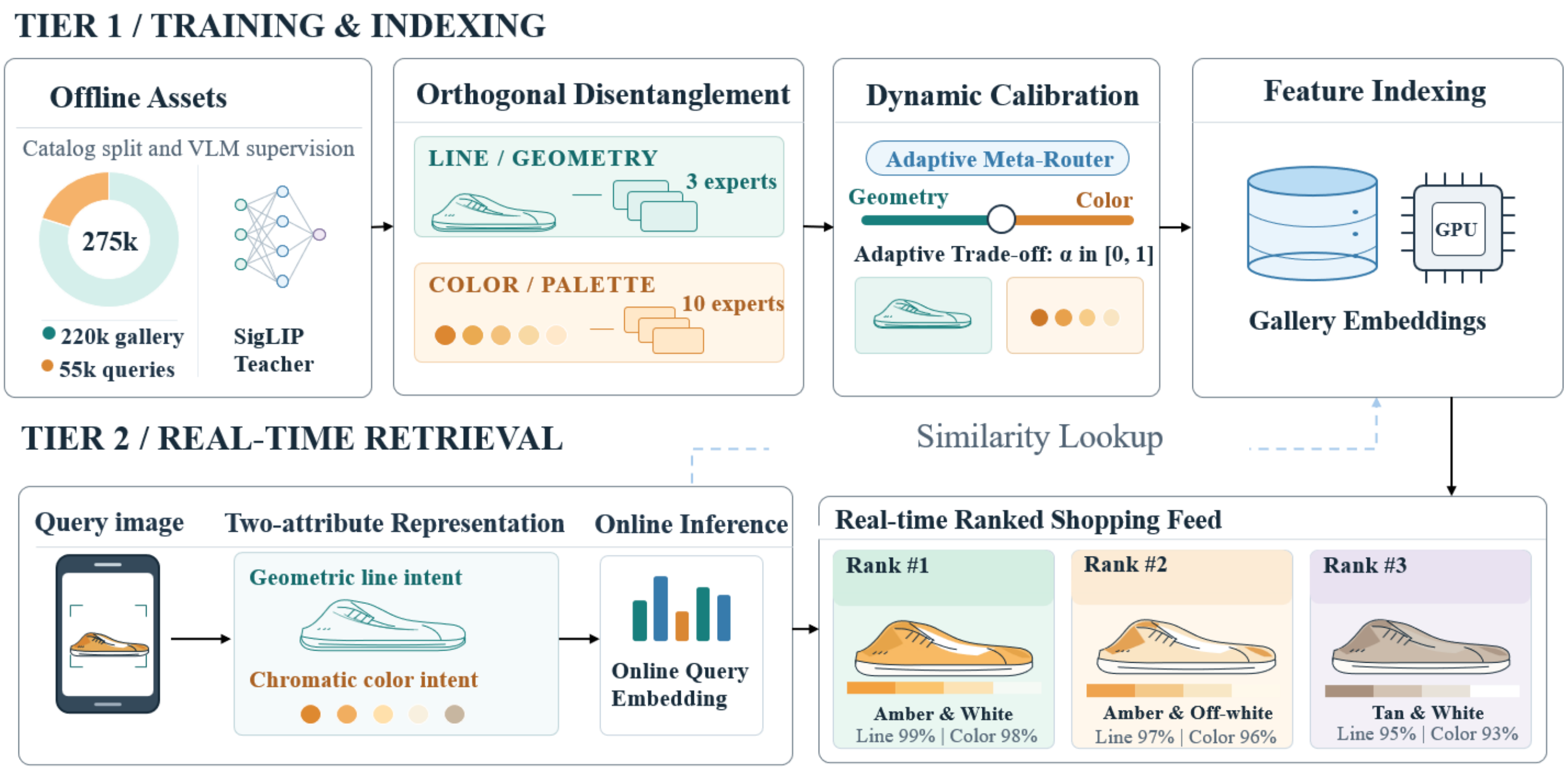

Fig. 1. Data partitioning, two-phase training, indexing, and online product retrieval workflow of PRISM-Net.

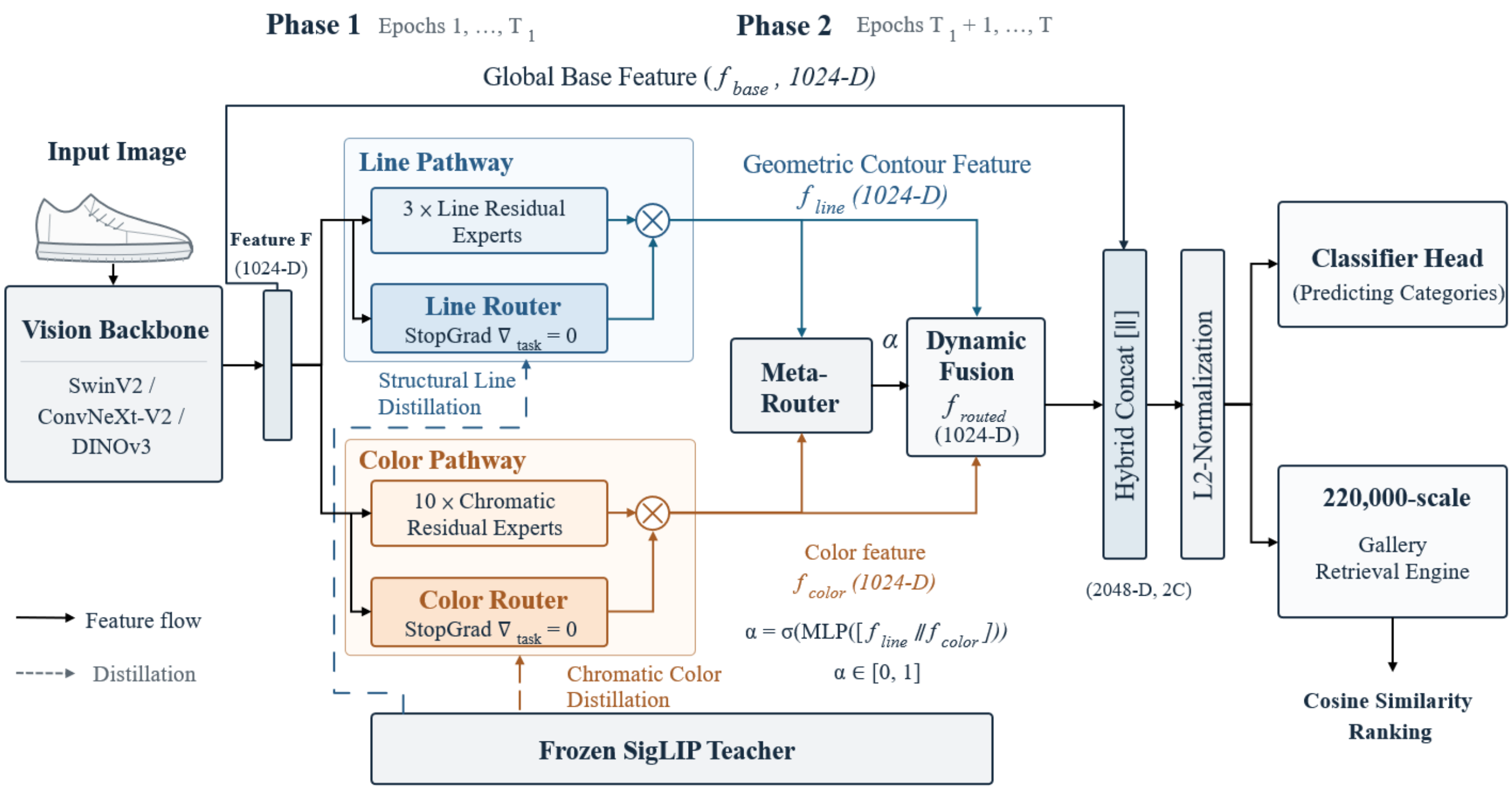

Fig. 2. Architecture and dual-purpose classification–retrieval pipeline of PRISM-Net.

## IV. RESULT

All experiments were conducted on an NVIDIA GeForce RTX 5090 (32 GB GDDR7 VRAM) GPU using PyTorch. To ensure strict reproducibility, 5-fold cross-validation splits were generated with a fixed random seed (42). Models were trained with a mini-batch size of 64 using the AdamW optimizer (weight decay 0.05) under a cosine annealing learning rate schedule (minimum learning rate $1 \times 10^{-6}$ ) and global gradient norm clipping at 1.0. The routing heads and multi-expert branches were initialized with a base learning rate of $\eta_{\text{head}} = 1 \times 10^{-3}$.

Backbone-specific tuning strategies were tailored to individual inductive biases:

- For SwinV2-Base, spatial feature maps were permuted to channel-first format to interface with convolutional experts, trained with a conservative backbone learning rate ( $\eta_{\text{bb}} = 1 \times 10^{-5}$ ) and identical heavy data augmentations to prevent self-attention divergence.
- For ConvNeXt-V2-Base, layer-wise learning rate Decay (LLRD, decay factor 0.875, backbone learning rate $\eta_{\text{bb}} = 5 \times 10^{-5}$ ), RandAugment, and Random Erasing were applied to stabilize deep pure-convolutional training.

- For DINOv3-Base, low-rank adaptation (LoRA, rank $r = 16$, scaling $\alpha = 16$, dropout 0.05) was injected into the query-key-value ( $qkv$ ) projections alongside lightweight cropping and flipping, preserving pre-trained self-supervised spatial patch representations while optimizing the dynamic routing mechanism.

### 1) Product Classification Accuracy

Table II compares the base, line-only, color-only, and PRISM-Net configurations across SwinV2, ConvNeXt-V2, and DINOv3. For SwinV2, line-only achieves the highest micro-accuracy (87.60%) and head-class accuracy (94.10%), improving them by 0.97 and 0.69 percentage points over base, respectively. Color-only performs best on macro-accuracy (72.04%), medium-class accuracy (59.91%), and tail-class accuracy (62.24%), corresponding to gains of 2.15, 2.29, and 3.60 points. PRISM-Net provides comparable improvements, reaching 87.57% micro-accuracy and 62.08% tail-class accuracy. For ConvNeXt-V2, line-only is consistently strongest, increasing micro-, macro-, medium-, and tail-class accuracy by 0.84, 2.79, 4.26, and 3.82 points over base, respectively, while increasing inference latency only slightly from 0.24 to 0.26 ms. For DINOv3, the base configuration retains the highest micro-accuracy (89.24%) and head-class accuracy (95.28%), whereas color-only yields the strongest macro-, medium-, and tail-class results at 76.58%, 65.31%, and 69.70%, representing gains of 1.53, 2.37, and 2.78 points. Overall, line-oriented routing is most beneficial for ConvNeXt-V2, while color-oriented routing provides the clearest gains for long-tail and class-balanced performance, particularly with SwinV2 and DINOv3.

TABLE II
CLASSIFICATION RESULTS OF FOUR ROUTING CONFIGURATIONS FOR DIFFERENT BACKBONES

| Backbone | Configuration | Micro Acc (Mean ± Std) | Macro Acc | Head (6) | Medium (6) | Tail (6) | Params (M) | Latency (ms) |
|---|---|---|---|---|---|---|---|---|
| SwinV2 | Base | 86.63% ± 0.14% | 69.89% | 93.41% | 57.62% | 58.64% | 87.4M | 0.34 |
| SwinV2 | Line-only | 87.60% ± 0.14% | 71.89% | 94.10% | 59.36% | 62.20% | 144.1M | 0.35 |
| SwinV2 | Color-only | 87.57% ± 0.14% | 72.04% | 93.97% | 59.91% | 62.24% | 276.4M | 0.38 |
| SwinV2 | PRISM-Net | 87.57% ± 0.12% | 71.85% | 94.05% | 59.42% | 62.08% | 333.2M | 0.40 |
| ConvNeXt-V2 | Base | 87.20% ± 0.15% | 70.66% | 93.91% | 58.18% | 59.90% | 88.2M | 0.24 |
| ConvNeXt-V2 | Line-only | 88.04% ± 0.14% | 73.45% | 94.20% | 62.44% | 63.72% | 144.9M | 0.26 |
| ConvNeXt-V2 | Color-only | 87.57% ± 0.59% | 71.83% | 93.92% | 60.36% | 61.21% | 277.1M | 0.30 |
| ConvNeXt-V2 | PRISM-Net | 87.77% ± 0.15% | 72.82% | 94.00% | 61.84% | 62.62% | 334.0M | 0.32 |
| DINOv3 | Base | 89.24% ± 0.19% | 75.05% | 95.28% | 62.94% | 66.92% | 86.0M | 0.37 |
| DINOv3 | Line-only | 88.98% ± 0.23% | 75.06% | 94.97% | 63.37% | 66.85% | 117.9M | 0.45 |
| DINOv3 | Color-only | 89.09% ± 0.13% | 76.58% | 94.73% | 65.31% | 69.70% | 192.3M | 0.62 |
| DINOv3 | PRISM-Net | 88.94% ± 0.25% | 74.89% | 94.94% | 63.44% | 66.30% | 224.3M | 0.70 |

### 2) Fine-grained Product Retrieval

The product retrieval performance of the base, line-only, color-only, and PRISM-Net configurations is evaluated using the entire validation split (55,091 unseen queries) against the training gallery (220,363 candidate images). Table III reports product mAP@10, soft-aligned line mAP@10, soft-aligned color mAP@10, gallery extraction time, and single-query search latency for each backbone and configuration.

As shown in Table III, the effects of attribute routing vary with the backbone. For SwinV2, line-only improves product, line, and color mAP@10 from 80.98%, 56.53%, and 26.06% to 83.84%, 62.27%, and 33.88%, corresponding to gains of 2.86, 5.74, and 7.82 percentage points. color-only provides the strongest SwinV2 results, achieving 83.90% product mAP@10 and 35.18% color mAP@10, with gains of 2.92 and 9.12 points over base; PRISM-Net reaches 83.42%, 57.08%, and 31.43% on the three retrieval metrics. For ConvNeXt-V2, base retains the highest product mAP@10 at 87.67%, whereas line-only provides a 0.68-point gain in line mAP@10. PRISM-Net achieves the highest line and color mAP@10 values of 59.95% and 35.87%, representing improvements of 0.69 and 1.55 points over base, although its product mAP@10 decreases by 0.83 points. For DINOv3, base produces the highest line mAP@10 (64.41%), while line-only and color-only improve product mAP@10 by 1.18 and 1.51 points, respectively. color-only also achieves the highest DINOv3 product and color mAP@10 values at 84.02% and 32.39%, although its line mAP@10 decreases by 0.93 points. Across all 12 models, gallery extraction requires 130.2–256.5 s, and cosine-similarity search over 220,363 candidates takes 0.005–0.010 ms per query. Overall, routing produces substantial and consistent gains for SwinV2, while ConvNeXt-V2 and DINOv3 exhibit clearer trade-offs between category-level product retrieval and attribute-level alignment.

TABLE III
RETRIEVAL RESULTS OF FOUR ROUTING CONFIGURATIONS FOR DIFFERENT BACKBONES

| Backbone | Model Configuration | Product mAP@10 | Line mAP@10 (Soft) | Color mAP@10 (Soft) | Gallery Time (s) | Latency (ms/Q) |
|---|---|---|---|---|---|---|
| SWINV2 | Base | 80.98% | 56.53% | 26.06% | 135.5s | 0.006 ms |
| SWINV2 | Line-Only | 83.84% | 62.27% | 33.88% | 138.0s | 0.010 ms |
| SWINV2 | Color-Only | 83.90% | 62.08% | 35.18% | 144.0s | 0.010 ms |
| SWINV2 | PRISM-Net | 83.42% | 57.08% | 31.43% | 148.0s | 0.010 ms |
| CONVNEXTV2 | Base | 87.67% | 59.26% | 34.32% | 130.2s | 0.006 ms |
| CONVNEXTV2 | Line-Only | 86.97% | 59.94% | 33.46% | 134.5s | 0.010 ms |
| CONVNEXTV2 | Color-Only | 86.82% | 59.77% | 35.70% | 143.5s | 0.010 ms |
| CONVNEXTV2 | PRISM-Net | 86.84% | 59.95% | 35.87% | 148.3s | 0.010 ms |
| DINOV3 | Base | 82.51% | 64.41% | 30.91% | 185.0s | 0.005 ms |
| DINOV3 | Line-Only | 83.69% | 63.44% | 32.00% | 201.5s | 0.008 ms |
| DINOV3 | Color-Only | 84.02% | 63.48% | 32.39% | 240.0s | 0.008 ms |
| DINOV3 | PRISM-Net | 82.40% | 63.96% | 32.22% | 256.5s | 0.008 ms |

### 3) Multi-Scale Retrieval Depth and Feature Manifold Analysis

Note that the base and PRISM-Net configurations are reported for SwinV2 and ConvNeXt-V2. For DINOv3, the representative single-attribute configuration (color-only) is additionally benchmarked to examine the cross-depth behavior of the teacher–student capacity inversion observed in Table III. This configuration avoids simultaneous multi-expert representation conflict. To examine ranking fidelity across varying retrieval depths, macro-balanced queries are evaluated (1,000 stratified queries across all 18 classes) against the 220,363-image gallery at depths K ∈ {1, 5, 10, 20, 50}. Table IV summarizes mAP at the five evaluated depths together with NDCG@10, color mAP@10, the Fisher inter-/intra-class discriminant ratio, and the cosine silhouette score. On SwinV2, PRISM-Net outperforms base at every evaluated depth, with the mAP margin increasing from +1.40 points at K = 1 (67.30% vs. 65.90%) to +4.16 points at K = 10 (64.97% vs. 60.81%) and +4.67 points at K = 50 (64.17% vs. 59.50%). Color mAP@10 also increases by 3.31 points (23.42% vs. 20.11%). In contrast, ConvNeXt-V2 PRISM-Net yields lower product mAP than base across all depths, including a 1.31-point decrease at K = 10 (70.91% vs. 72.22%), although color mAP@10 increases by 1.22 points (27.40% vs. 26.18%). On DINOv3, PRISM-Net remains close to base, whereas color-only achieves the strongest cross-depth results: mAP@1 reaches 76.30%, mAP@50 improves by 6.22 points (61.96% vs. 55.74%), NDCG@10 improves by 2.99 points (73.73% vs. 70.74%), and the cosine silhouette score increases from 0.015 to 0.058 (3.87×). Fig. 3 visualizes the corresponding feature manifolds, illustrating the backbone- and configuration-dependent changes summarized by Table IV.

TABLE IV
MULTI-SCALE RETRIEVAL EVALUATION ACROSS DEPTHS K IN {1, 5, 10, 20, 50}, NDCG@10, FISHER RATIO, AND COSINE SILHOUETTE

| **Backbone** | **Model Name** | **mAP@1** | **mAP@5** | **mAP@10** | **mAP@20** | **mAP@50** | **NDCG@10** | **Color@10** | **Fisher Ratio** | **Silhouette** |
|---|---|---|---|---|---|---|---|---|---|---|
| SWINV2 | Base | 65.90% | 61.71% | 60.81% | 60.04% | 59.50% | 65.00% | 20.11% | 1.002 | 0.223 |
| SWINV2 | PRISM-Net | 67.30% | 65.30% | 64.97% | 64.46% | 64.17% | 67.36% | 23.42% | 0.945 | 0.236 |
| CONVNEXTV2 | Base | 72.60% | 72.30% | 72.22% | 72.18% | 72.11% | 72.80% | 26.18% | 0.612 | 0.240 |
| CONVNEXTV2 | PRISM-Net | 72.50% | 71.18% | 70.91% | 70.57% | 70.11% | 72.26% | 27.40% | 0.441 | 0.156 |
| DINOV3 | Base | 75.20% | 68.18% | 64.62% | 60.83% | 55.74% | 70.74% | 25.07% | 0.167 | 0.015 |
| DINOV3 | PRISM-Net | 75.20% | 68.45% | 64.72% | 60.93% | 55.73% | 70.88% | 25.79% | 0.171 | 0.027 |
| DINOV3 | Color-Only | 76.30% | 71.10% | 68.33% | 65.48% | 61.96% | 73.73% | 25.83% | 0.197 | 0.058 |

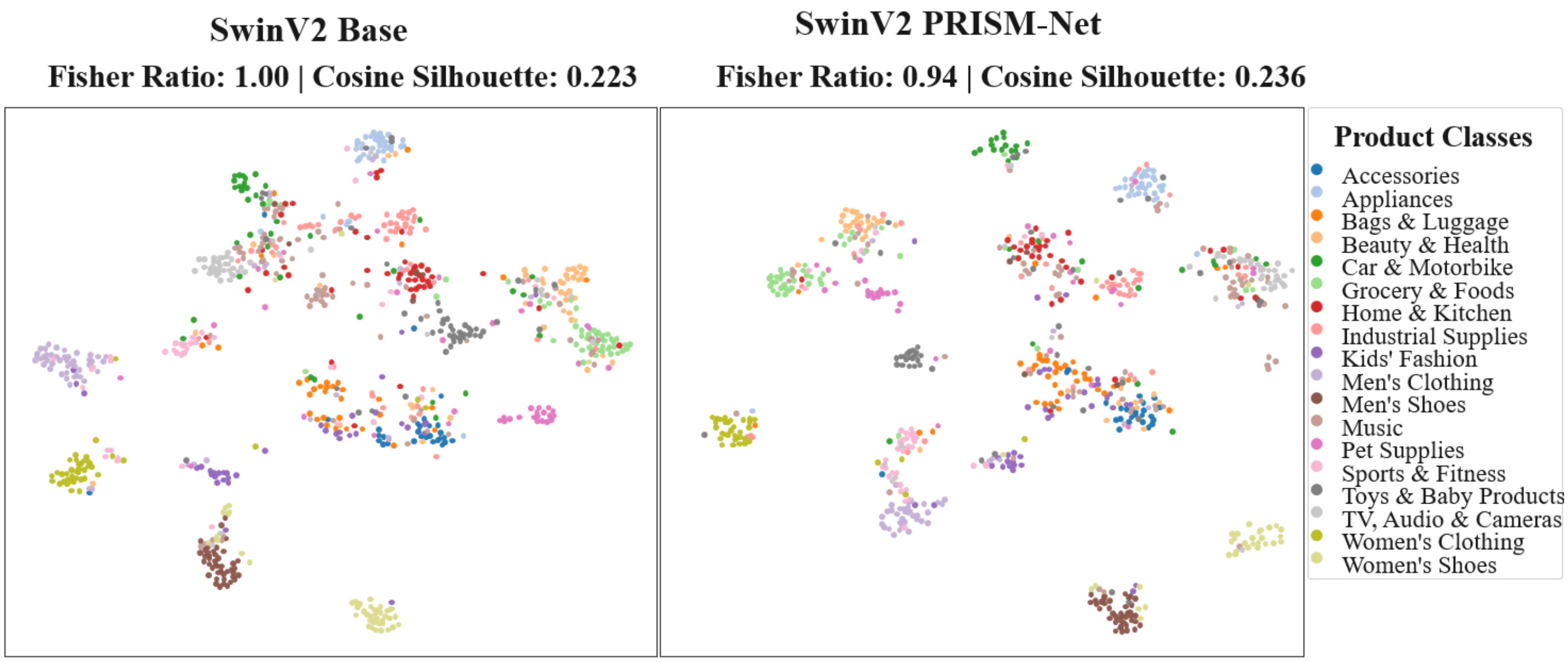

Fig. 3. 2D t-SNE visualization [34] of high-dimensional retrieval feature manifolds on the 18-class test set comparing Base (left) and PRISM-Net (right), annotated with Fisher discriminant ratios and Cosine Silhouette scores.

#### 4) Unsupervised Semantic Interpretability of the Meta-Router Gate

Fig. 4 show that PRISM-Net's meta-router adapts line and color expert weights to product characteristics without category-level supervision. Across both DINOv3 and ConvNeXt-V2, rigid, contour-defined categories receive higher line weights, whereas clothing and other color-dominant categories allocate over 95% of their weight to the color expert. For example, DINOv3 assigns Appliances a 6.89x higher line weight than Women's Clothing. These consistent results demonstrate explainable, category-adaptive dynamic routing across both ViT and CNN backbones.

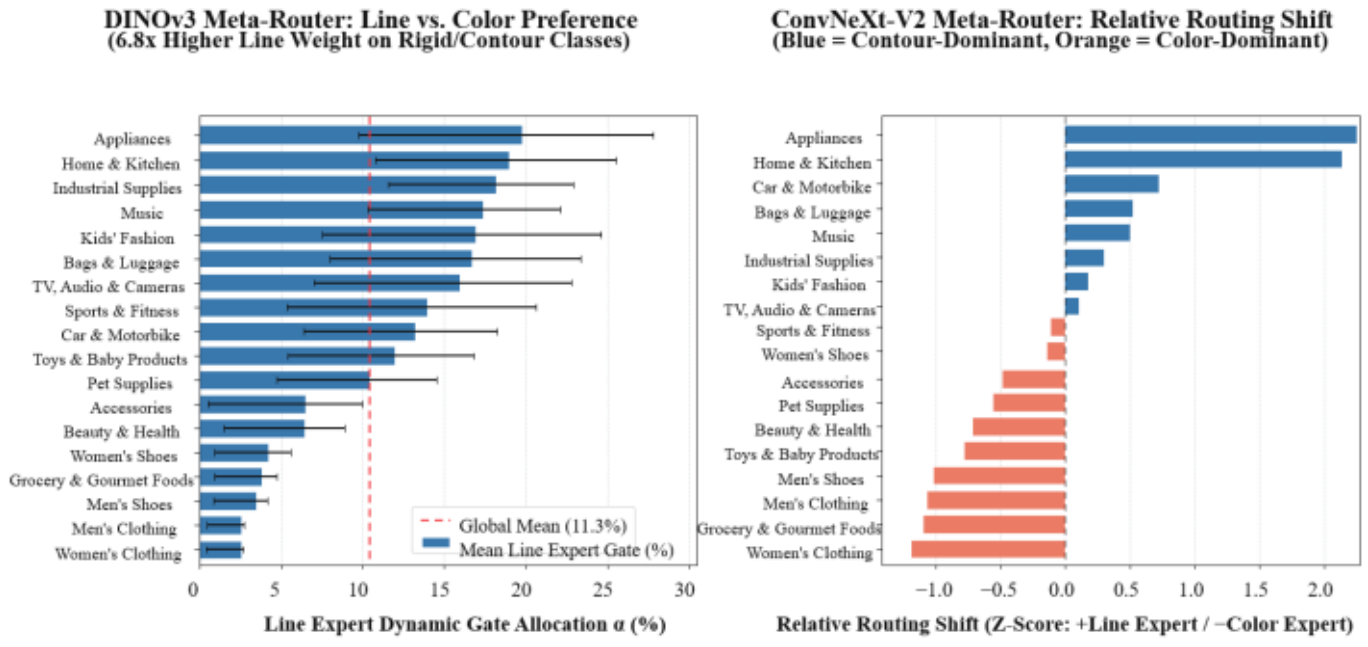

Fig. 4. Cross-category interpretability analysis of the conditional meta-router gate alpha across all 18 product categories (55,091 test images): (Left) DINOv3 line expert gate allocation alpha (%), showing a 6.8x higher structural weight on rigid appliances/hardware than on apparel; (Right) ConvNeXt-V2 relative routing shift (Z-score), demonstrating identical physical category preferences across distinct backbone families.

#### 5) Qualitative Attribute-Guided Retrieval

Figs. 5–7 qualitatively demonstrate how PRISM-Net disentangles structural and chromatic attributes in Top-10 retrieval. For Appliances, the line expert eliminates base-model confusion with audio equipment and achieves a 10/10 category hit rate. For Women's Clothing, the color expert prevents chromatic drift, improving color similarity from 0.220 to at least 0.99. For Pet Supplies, PRISM-Net avoids confusion between metal dog collars and jewelry, again achieving 10/10 correct-category retrieval. These examples confirm that the specialized experts improve category accuracy while preserving the query's key visual attributes.

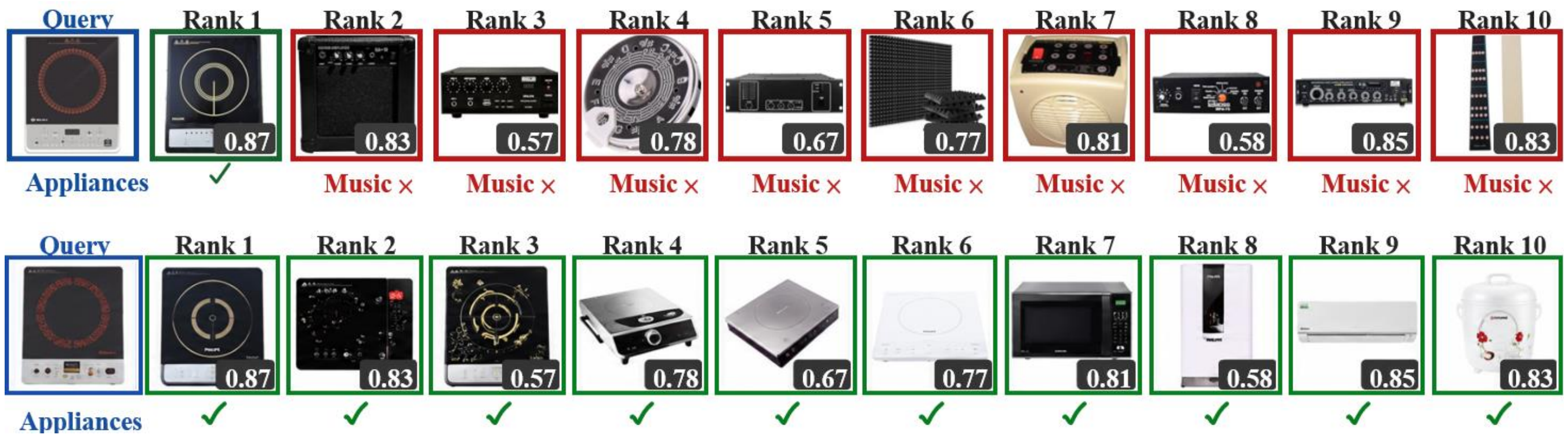

Fig. 5. Top-10 retrieval results for a structural line-focused Appliances query: (a) the base confuses control buttons and the dark chassis with musical equipment, producing 9/10 false positives (red borders); (b) PRISM-Net achieves 10/10 category hits (green borders) through disentangled concentric-coil contour alignment.

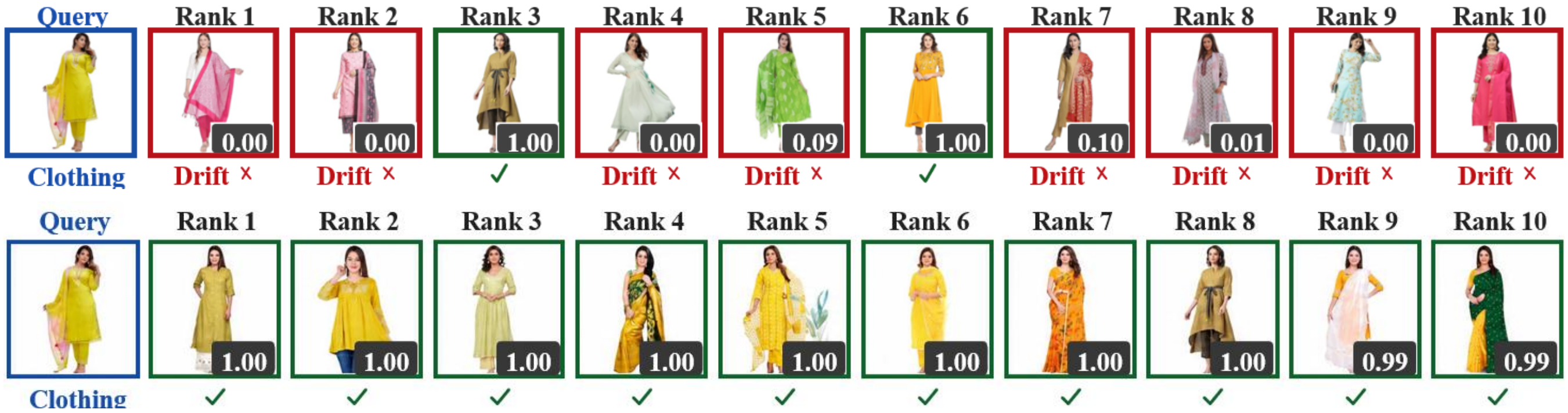

Fig. 6. Top-10 retrieval results for a color-focused Women's Clothing query: (a) the base exhibits severe chromatic drift, retrieving pink, green, and white items (mean C_Sim = 0.220); (b) PRISM-Net preserves chromatic consistency across all top-ranked items (C_Sim >= 0.99).

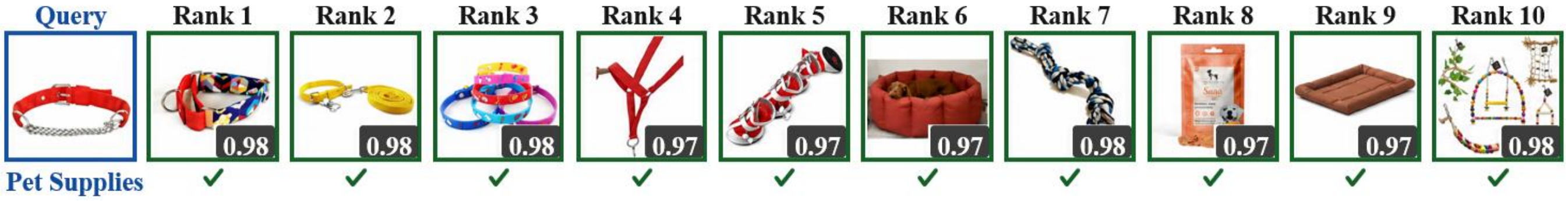

Fig. 7. PRISM-Net qualitative Top-10 retrieval showcase on the high-gain category (Pet Supplies, +5.18% mAP), demonstrating robust cross-attribute disambiguation (10/10 category hits) compared to 9/10 jewelry false-positive errors for the base.

## V. DISCUSSION

Table II shows that increased parameter capacity does not produce uniform gains across backbones or routing configurations. PRISM-Net improves medium- and tail-class performance over base on ConvNeXt-V2 and SwinV2, but the larger DINOv3 full model does not exceed the base configuration in overall micro- or macro-accuracy. In contrast, the specialist variants often provide the strongest balanced results: ConvNeXt-V2 line-only reaches 73.45% macro-accuracy and 63.72% tail-class accuracy, while DINOv3 color-only reaches 76.58% and 69.70%, respectively. These comparisons indicate that the gains depend primarily on attribute-specific routing and backbone compatibility rather than parameter count alone.

Comparing the three backbone families reveals a profound theoretical dichotomy. First, SwinV2 shares homogeneous patch-token attention priors with the ViT-based SigLIP teachers, enabling frictionless cross-modal transfer (+2.92% product mAP, +9.12% color mAP). Second, ConvNeXt-V2 extracts strictly translation-equivariant local features via 7x7 depthwise convolutions; forcing its intermediate feature maps to simultaneously satisfy 13 global text-aligned VLM prompts introduces inductive bias friction, slightly trading off macro head-class accuracy (-0.33%) in exchange for enhanced chromatic retrieval (+1.55% color mAP). Notably, relaxing the constraint to a single structural dimension (ConvNeXt-V2 line-only) eliminates this friction, restoring 88.04% micro-accuracy and achieving peak 73.45% macro-accuracy.

Third, DINOv3 exhibits a phenomenon termed teacher-student capacity inversion. Trained via large-scale self-supervised discriminative masked modeling, DINOv3 possesses native pixel-level edge sharpness (line mAP@10 = 64.41% out of the box) that surpasses the coarse 3-prompt linguistic resolution of SigLIP. Consequently, imposing joint dual-branch distillation slightly bottlenecks DINOv3's native structural features (82.40% mAP). However, when decoupled into a single specialist branch (DINOv3 color-only or line-only), the routing module complements DINOv3's geometry with chromatic awareness without over-constraining its feature space, unlocking a massive +1.51% Product mAP surge (84.02%) and a +5.03% leap on hard long-tail categories.

Beyond backbone-dependent performance, the learned meta-router exhibits emergent physical interpretability. Category-level gate analysis shows that both ViT and CNN backbones assign up to 6.8× higher line expert weight to rigid, contour-driven categories such as appliances, kitchenware, and hardware, while allocating more than 96% color expert weight to appearance-driven categories such as apparel, footwear, and groceries. These routing patterns were not imposed through manual attribute labels; rather, they emerged from the attribute-specific VLM-guided objectives. This result suggests that PRISM-Net's performance gains are accompanied by semantically meaningful expert specialization, while also indicating that routing behavior should be interpreted relative to category appearance and backbone inductive bias.

Unlike the mutually exclusive head, medium, and tail tiers defined by sample frequency, the hard subset is a baseline-dependent diagnostic group. It comprises the six categories whose base classification accuracy is below 65%, Bags & Luggage, Music, Industrial Supplies, Pet Supplies, Sports & Fitness, and Kids' Fashion, and therefore overlaps the frequency tiers. Because both the threshold and membership depend on the reference model, hard-class accuracy is excluded from the primary comparison in Table II and interpreted only as supplementary evidence. Under this diagnostic, the strongest specialist configurations improve accuracy over base by 5.31 points on ConvNeXt-V2, 4.36 points on DINOv3, and 3.23 points on SwinV2, supporting the conclusion that attribute routing is particularly useful for visually heterogeneous categories

## VI. CONCLUSION

This paper presents PRISM-Net, an explainable multi-expert framework for fine-grained multimedia product classification and retrieval. By coupling zero-shot VLM sigmoid semantic distillation with stop-gradient expert routing and an instance-adaptive meta-router gate alpha, PRISM-Net effectively resolves shape-color feature entanglement without manual attribute annotations. Exhaustive 5-Fold cross-validation on 275,454 images and full-scale retrieval evaluations across 55,091 queries and 220,363 gallery items across SwinV2, ConvNeXt-V2, and DINOv3 validate the effectiveness, sub-millisecond retrieval speed (0.01 ms/query), and emergent physical interpretability of our approach. Future work will explore multi-teacher spatial-semantic fusion, combining DINOv3 spatial maps with multimodal long-context and mixture-of-vision-expert reasoning priors, and structural re-parameterization to compress multi-expert branches into a zero-overhead single-branch inference network.